\documentclass[conference]{IEEEtran}
\IEEEoverridecommandlockouts
\usepackage{cite}
\usepackage{amsmath,amssymb,amsfonts}
\usepackage{algorithmic}
\usepackage{graphicx}
\usepackage{booktabs}
\usepackage{float}
\usepackage{multirow}  
\usepackage{textcomp}
\usepackage{xcolor}
\def\BibTeX{{\rm B\kern-.05em{\sc i\kern-.025em b}\kern-.08em
    T\kern-.1667em\lower.7ex\hbox{E}\kern-.125emX}}
\begin{document}

\title{FAST: A Synergistic Framework of Attention and State-space Models for Spatiotemporal Traffic Prediction
}

\newif\ifarxiv
\arxivtrue  

\author{
\IEEEauthorblockN{Xinjin Li$^{*}$\thanks{* Equal contribution.}}
\IEEEauthorblockA{
\textit{Columbia University} \\
New York, NY, USA \\
li.xinjin@columbia.edu}
\and
\IEEEauthorblockN{Jinghan Cao$^{*}$}
\IEEEauthorblockA{
\textit{San Francisco State University} \\
San Francisco, CA, USA \\
jcao3@alumni.sfsu.edu}
\and
\IEEEauthorblockN{Mengyue Wang}
\IEEEauthorblockA{
\textit{University of California, Berkeley} \\
Berkeley, CA, USA \\
cwang998@berkeley.edu}
\and
\IEEEauthorblockN{Yue Wu}
\IEEEauthorblockA{
\textit{New York University} \\
New York, NY, USA \\
y493423@gmail.com}
\and
\IEEEauthorblockN{Longxiang Yan}
\IEEEauthorblockA{
\textit{University of Pennsylvania} \\
Philadelphia, PA, USA \\
tonyyan@alumni.upenn.edu}
\and
\IEEEauthorblockN{Yeyang Zhou}
\IEEEauthorblockA{
\textit{University of California, San Diego} \\
La Jolla, CA, USA \\
yeyang-zhou@ucsd.edu}
\and
\IEEEauthorblockN{Ziqi Sha}
\IEEEauthorblockA{
\textit{University of Massachusetts Amherst} \\
Amherst, MA, USA \\
ziqisha@umass.edu}
\and
\IEEEauthorblockN{Yu Ma}
\IEEEauthorblockA{
\textit{Carnegie Mellon University} \\
Pittsburgh, PA, USA \\
yuma13926@gmail.com}
}
\maketitle

\begin{abstract}
Traffic forecasting requires modeling complex temporal dynamics and long-range spatial dependencies over large sensor networks. Existing methods typically face a trade-off between expressiveness and efficiency: Transformer-based models capture global dependencies well but suffer from quadratic complexity, while recent selective state-space models are computationally efficient yet less effective at modeling spatial interactions in graph-structured traffic data. We propose FAST, a unified framework that combines attention and state-space modeling for scalable spatiotemporal traffic forecasting. FAST adopts a Temporal-Spatial-Temporal architecture, where temporal attention modules capture both short- and long-term temporal patterns, and a Mamba-based spatial module models long-range inter-sensor dependencies with linear complexity. To better represent heterogeneous traffic contexts, FAST further introduces a learnable multi-source spatiotemporal embedding that integrates historical traffic flow, temporal context, and node-level information, together with a multi-level skip prediction mechanism for hierarchical feature fusion. Experiments on PeMS04, PeMS07, and PeMS08 show that FAST consistently outperforms strong baselines from Transformer-, GNN-, attention-, and Mamba-based families. In particular, FAST achieves the best MAE and RMSE on all three benchmarks, with up to 4.3\% lower RMSE and 2.8\% lower MAE than the strongest baseline, demonstrating a favorable balance between accuracy, scalability, and generalization.

\end{abstract}


\section{Introduction}

\begin{figure}[t]
    \centering
    \includegraphics[width=\linewidth]{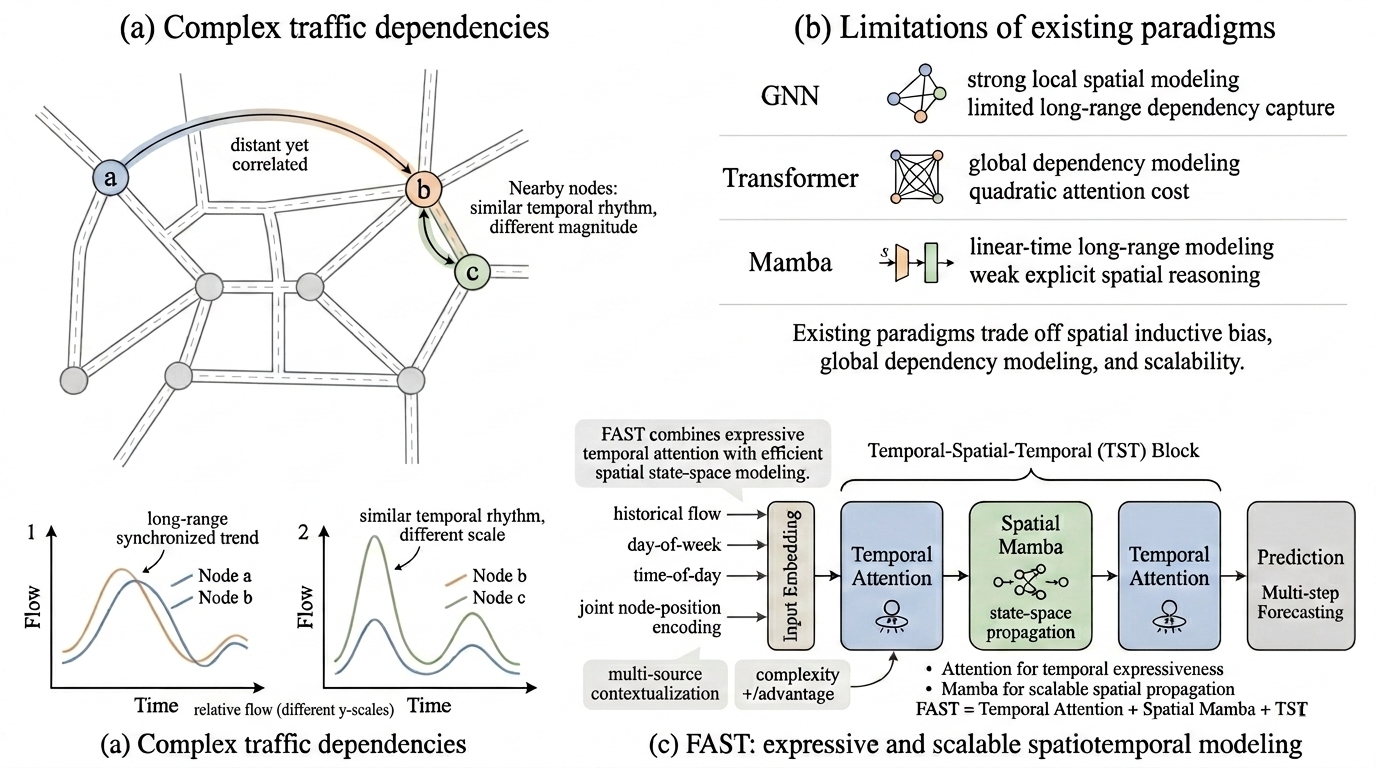}
    \caption{Motivation of FAST. (a) Traffic forecasting exhibits complex spatiotemporal dependencies. (b) Existing paradigms trade off local spatial bias, global dependency modeling, and scalability. (c) FAST combines temporal attention and Mamba-based spatial propagation in a TST design.}
    \label{fig:motivation}
\end{figure}

Spatiotemporal traffic forecasting is a core problem in intelligent transportation systems, where the goal is to predict future traffic states over large sensor networks from historical observations. Effective forecasting requires a model to simultaneously capture long-range temporal dependencies, non-local spatial correlations, and computational scalability. This remains challenging because traffic interactions are neither purely local nor static. As illustrated in Fig.~\ref{fig:motivation}(a), geographically distant sensors may exhibit highly synchronized dynamics, while nearby sensors can share similar temporal rhythms but differ substantially in magnitude. Therefore, accurate traffic forecasting calls for a spatiotemporal architecture that is expressive in time, effective in modeling complex spatial interactions, and scalable to realistic large-scale deployments.

Existing paradigms each address only part of this design space. Graph neural network (GNN) based methods have become a standard choice for traffic forecasting because they exploit road-network structure to model spatial correlations. However, their spatial inductive bias is often dominated by local message passing, which limits the ability to capture long-range dependencies among functionally related but distant nodes. Attention-based models alleviate this locality bias by flexibly modeling global dependencies and have shown strong performance in time-series forecasting. Yet their quadratic complexity makes them increasingly expensive for long historical windows and large sensor networks. More recently, selective state-space models (SSMs), such as Mamba, have emerged as an efficient alternative for long-range dependency modeling due to their linear-time complexity. Nevertheless, these models are primarily designed for sequential dependency propagation and are less naturally aligned with the structured spatial reasoning required in traffic forecasting. As summarized in Fig.~\ref{fig:motivation}(b), existing methods still face a persistent trade-off among temporal expressiveness, spatial effectiveness, and scalability.

Our key insight is that temporal and spatial dependencies impose fundamentally different modeling requirements. Temporal evolution benefits from attention, which can flexibly relate distant observations and capture both short-term fluctuations and long-term temporal patterns through fine-grained contextual weighting. Spatial propagation across sensors, in contrast, can be modeled more efficiently through selective state-space transitions, which are well suited for long-range information propagation with linear complexity. This complementarity suggests a natural division of labor: use attention for temporal reasoning and use a Mamba-based state-space module for scalable spatial propagation.

Based on this insight, we propose \textbf{FAST}, a \textbf{F}ramework of \textbf{A}ttention and \textbf{S}tate-space for spatiotemporal \textbf{T}raffic prediction. As shown in Fig.~\ref{fig:motivation}(c), FAST adopts a \emph{Temporal-Spatial-Temporal} (TST) architecture, in which a spatial Mamba module is interleaved between two temporal attention modules. The first temporal module extracts expressive node-wise temporal representations, the spatial module propagates information across sensors to capture long-range inter-sensor dependencies, and the second temporal module refines temporal dynamics under the updated spatial context. This interleaved design allows temporal reasoning and spatial propagation to iteratively reinforce one another, yielding a unified and scalable spatiotemporal representation.

Beyond the core TST design, FAST further incorporates a learnable multi-source spatiotemporal embedding that integrates historical traffic observations, time-of-day, day-of-week, and node-specific positional information, enabling the model to capture both periodic patterns and heterogeneous sensor behaviors. In addition, FAST employs a multi-level skip prediction mechanism to aggregate hierarchical representations across stacked blocks, improving multi-scale feature fusion and stabilizing multi-step forecasting.

Extensive experiments on three widely used real-world benchmarks, PeMS04, PeMS07, and PeMS08, demonstrate the effectiveness of FAST. Compared with strong baselines from GNN-, attention-, Transformer-, and Mamba-based families, FAST achieves the best MAE and RMSE on all three datasets while maintaining competitive MAPE. In particular, on PeMS08, FAST reduces RMSE by up to 4.3\% and MAE by up to 2.8\% relative to the strongest baseline. These results show that FAST achieves a favorable balance among accuracy, scalability, and generalization.

Our main contributions are summarized as follows:
\begin{itemize}
    \item We identify a key design gap in spatiotemporal traffic forecasting: existing GNN-, attention-, and state-space-based methods each capture only part of the trade-off among temporal expressiveness, long-range spatial interaction modeling, and computational scalability.
    \item We propose FAST, a unified spatiotemporal forecasting framework that combines temporal attention with Mamba-based spatial propagation through a novel Temporal-Spatial-Temporal architecture, together with learnable multi-source spatiotemporal embedding and hierarchical skip-connected prediction.
    \item We conduct extensive experiments on three real-world benchmarks and show that FAST consistently outperforms strong baselines, achieving the best MAE and RMSE on all three datasets while demonstrating strong robustness and scalability.
\end{itemize}

\section{Related Work}

Predicting traffic flow is fundamental for intelligent transportation systems, as it enables efficient traffic control, congestion alleviation, and dynamic resource scheduling. Designing accurate forecasting models remains challenging due to the stochastic and nonlinear nature of traffic dynamics. Beyond forecasting, reinforcement learning-based traffic management has also shown adaptability in complex transportation systems; for example, Duan et al. \cite{11059994} proposed a Bayesian critique-tune framework with adaptive pressure for multi-intersection traffic signal control, offering an effective solution for coordinated control under complex traffic conditions.

\ifarxiv

\subsection{Broader Context in Efficient Spatiotemporal Modeling.}
Beyond traditional traffic forecasting, the challenge of capturing complex spatiotemporal dynamics and multi-level graph dependencies extends across broader AI domains. Recent advances in spatiotemporal alignment, dynamic scene reconstruction, and multi-view continuous learning highlight the necessity of decoupling spatial and temporal priors to mitigate feature interference \cite{wu2025spatiotemporal, ning2025multi, yan2025turboreg, yu2026spatiotemporal, liu2024difflow3d, liu2024dvlo}. Notably, explicit spacetime modeling \cite{jiang2025stg} and adaptive dynamic representations \cite{chan2026adagar} provide highly effective priors for such structural transitions. To further manage intricate relational structures, researchers have increasingly relied on advanced graph neural networks, unified vector priors, and multi-graph architectures \cite{ouyang2024learn, ke2025stable, liu2025research, zhao2025efficient, zeng2026priordrive, jiang2026magma, hu2024decentralized}. Furthermore, fine-grained state control, hierarchical feature disambiguation, and tri-subspace disentanglement have proven critical for robust contextual understanding \cite{ji2025finestate, INTENT, HUD, REFINE, meng2026tri, yu2025cotextor, luo-etal-2025-dynamicner, codes2026, pesf-kd}. FAST draws inspiration from these representation paradigms, leveraging its Temporal-Spatial-Temporal (TST) architecture and multi-source embeddings to systematically disentangle and fuse complex urban mobility patterns.

Concurrently, as spatiotemporal models scale for industrial deployment, there is an industry-wide imperative to balance expressiveness with computational efficiency and robustness. The inherent quadratic constraints of standard transformers have motivated explorations into efficient sequence modeling, hardware co-design, and linear-time state-space alternatives \cite{lai2026transformers, liu2023regformer, chen2025autoneuralcodesigningvisionlanguagemodels, deng2026rearl, jiang2026anatomy}. For large-scale distributed systems and resource-constrained edge environments, recent frameworks heavily emphasize model compression, quantization, and deployment-friendly architectures \cite{liu2025efficient, liu2024fedlpa, liu2025one, zhou2023fastpillars, gsq, JUNJIE2025581, wang-etal-2025-filter, xiao2026echo, li2025efficient, wu2023semidefinite}. Notably, dynamic token compression \cite{anonymous2025comptrack} drastically reduces the computational overhead of such complex sequence tracking. Moreover, real-world deployment in safety-critical infrastructures—such as autonomous driving and intelligent transportation—requires models to exhibit strong adaptivity to regime-dependent volatility, resilience against noisy data, and provable safety guarantees \cite{cheng2026regime, wu2025efficiency, rao2025dynamicsamplingadaptsiterative, Feng_2024_NoiseBox, chen2025prosac, feng2025noisy, luan2025dynamic, luo2026agentauditorhumanlevelsafetysecurity, pang2026voice, hu2025reinforcement}. Ultimately, the high-fidelity predictions generated by these robust systems serve as a foundational layer for downstream tasks like realistic traffic signal control and sustainable urban planning \cite{li2025fully, zhang2025ccma, he2021towards, wang-etal-2025-reasoning-enhanced, yu2026rulesfallshortagentdriven, zeng2025futuresightdrive, zeng2025janusvln}. By integrating a linear-time Mamba spatial module, FAST directly addresses these scalability and robustness demands, ensuring efficient inference without sacrificing predictive capacity.
\fi

\subsection{Traditional Statistical and Machine Learning Methods.}
Early studies based on statistical and shallow learning models (e.g., Historical Average (HA) \cite{HA} and Autoregressive Integrated Moving Average (ARIMA) \cite{ARIMA}) captured short-term trends but failed on nonlinear multivariate traffic patterns. Flexible regressors such as Linear Regression (LR) \cite{LR} and Support Vector Regression (SVR) \cite{SVR} improved modeling accuracy in some cases, but still struggled to represent multivariate interactions and long-term dependencies.

\subsection{Deep Learning-Based Methods.}
With the growth of large-scale sensing data, deep neural networks have become dominant in traffic forecasting due to their ability to model nonlinear dynamics.

\emph{Graph-Based Methods.} Researchers incorporated graph convolutional networks (GCNs) to capture spatial correlations. DCRNN \cite{DCRNN}, STGCN \cite{STGCN}, GWNET \cite{GWNET}, STFGNN \cite{STFGNN}, and AGCRN \cite{AGCRN} combined graph convolutions with recurrent or temporal modules, adaptive adjacency learning, or node-level parameterization. However, these convolution-based methods remain inherently local, limiting their ability to model long-range dependencies among distant yet functionally related nodes.

\emph{Attention-Based Models.} Inspired by classic \cite{Transformer}, attention mechanisms enhance the capture of long-range dependencies. GMAN \cite{GMAN}, ST-WA \cite{ST-WA}, PatchTST \cite{PatchTST}, and iTransformer \cite{iTransformer} improve dynamic weighting, temporal aggregation, and cross-variable dependency modeling. In a related direction, Duan et al. \cite{duanadaptive} explored adaptive context length optimization for multi-agent reinforcement learning, providing a thoughtful perspective on balancing contextual modeling capacity and efficiency. However, the quadratic computational complexity of self-attention limits scalability on large traffic networks.

\subsection{State Space Model Methods.}
Many sequence models, including RNNs and Transformers, struggle with long-term dependency retention due to information decay. The HIPPO framework \cite{HIPPO} mitigated this issue by maintaining high-order polynomial projections, while the S4 model \cite{S4} introduced structured state-space parameterization and fast convolutions for efficient long-sequence modeling. However, static parameters limit adaptability in dynamic environments. The Mamba architecture \cite{Mamba} addressed this with input-dependent selective mechanisms that dynamically adjust state transitions, significantly improving precision and efficiency. Mamba has since been adapted to diverse domains \cite{U-Mamba, VIM, VideoMamba}, demonstrating strong generalization capacity. Despite their effectiveness, Mamba-style state space mechanisms remain underexplored for traffic forecasting.

Building upon these advances, our work integrates the complementary advantages of transformers and state-space models within a unified spatiotemporal architecture.
\section{Methodology}

\begin{figure}[t]
    \centering
    \includegraphics[width=\linewidth]{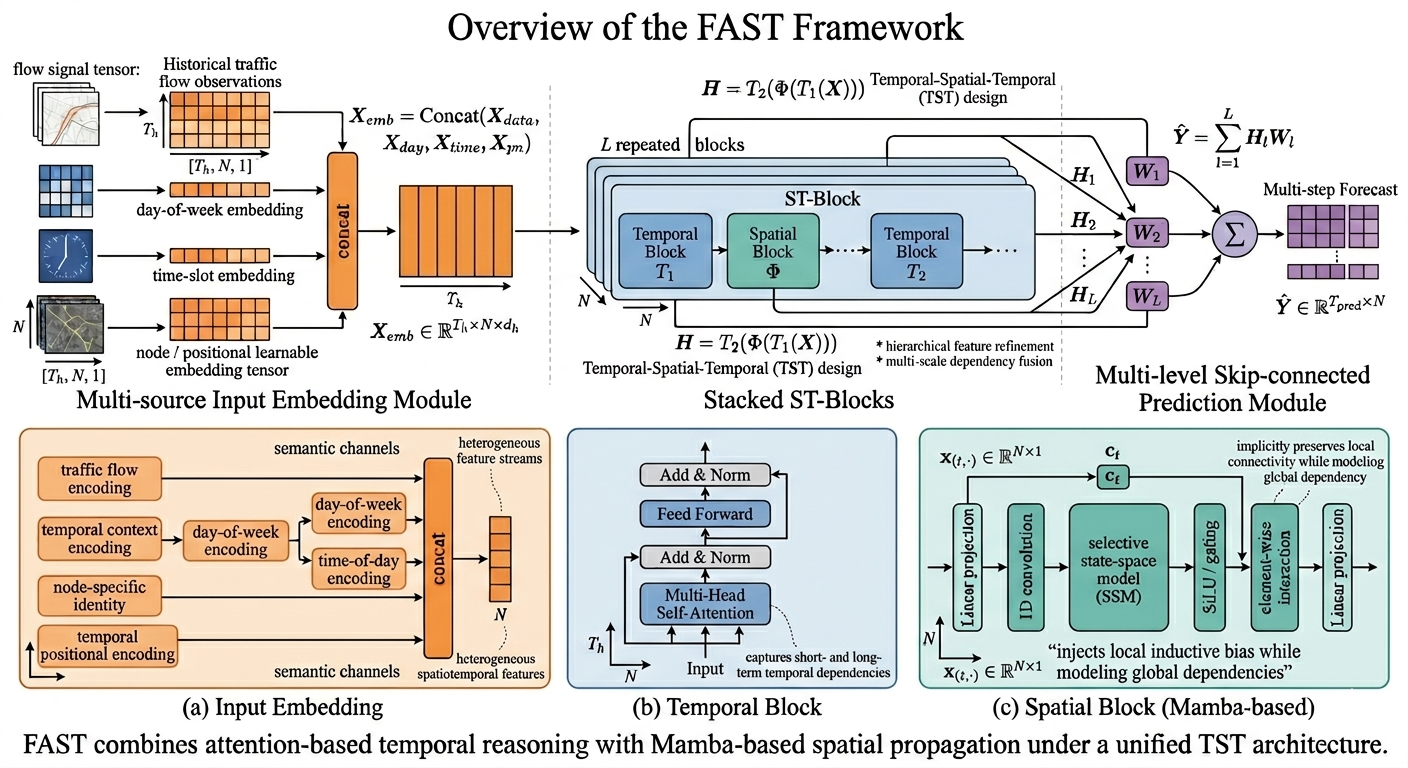}
    \caption{Overview of the FAST framework. FAST uses multi-source input embedding, stacked Temporal-Spatial-Temporal (TST) blocks, and a skip-connected prediction head for multi-step forecasting.}
    \label{fig:model}
\end{figure}

\noindent
We present \textbf{FAST}, a unified framework for spatiotemporal traffic forecasting that combines attention-based temporal reasoning with Mamba-based spatial propagation. This research was partially supported by Shanghai Elegant Technology Co., Ltd. As shown in Fig.~\ref{fig:model}, FAST first maps heterogeneous inputs into a shared latent space, then applies $L$ stacked \emph{Temporal-Spatial-Temporal} (TST) blocks, and finally aggregates multi-level features through a skip-connected prediction head for multi-step forecasting.

\subsection{Overview and Design Rationale}

FAST adopts a \emph{Temporal-Spatial-Temporal} (TST) design, using self-attention for temporal reasoning and selective state-space propagation for scalable inter-sensor modeling. In each block, a temporal attention module extracts node-wise temporal features, a Mamba-based spatial module propagates information across sensors, and a second temporal module refines temporal dynamics under the updated spatial context.

\subsection{Problem Setup}

Let $X_t \in \mathbb{R}^{N}$ denote the traffic flow observed at time step $t$ from $N$ sensors. Given the most recent $T_h$ observations, the input sequence is
$
X = [X_{t-T_h+1}, X_{t-T_h+2}, \ldots, X_t]^\top \in \mathbb{R}^{T_h \times N}.
$
The forecasting objective is to learn
$
f: \{X_{t-T_h+1}, \ldots, X_t\} \mapsto \{X_{t+1}, \ldots, X_{t+T_f}\},
$
where $T_f$ is the prediction horizon.

After input embedding, FAST produces
$
H^{(0)} = X_{\text{emb}} \in \mathbb{R}^{T_h \times N \times d_h},
$
which is processed by $L$ stacked TST blocks:
$
H^{(l)} = \mathcal{T}_2^{(l)}\!\left(\Phi^{(l)}\!\left(\mathcal{T}_1^{(l)}\!\left(H^{(l-1)}\right)\right)\right), \quad l=1,\ldots,L,
$
where $\mathcal{T}_1^{(l)}$ and $\mathcal{T}_2^{(l)}$ are temporal attention modules and $\Phi^{(l)}$ is the Mamba-based spatial propagation module.

\subsection{Temporal-Spatial-Temporal Block}

The TST block is FAST's core unit. Instead of loosely decoupling temporal and spatial modeling, FAST interleaves them so that temporal reasoning and spatial propagation iteratively refine each other. As in Fig.~\ref{fig:model}(b)--(c), each block contains two temporal modules surrounding one spatial module.

\subsubsection{Temporal Attention Module}

The temporal module models node-wise dependencies along the historical time dimension. For sensor $n$, let
$
H^{(l-1)}_{:n:} \in \mathbb{R}^{T_h \times d_h}
$
be its temporal feature sequence from the previous layer. We apply multi-head self-attention independently to each node:
$
Q_n = H^{(l-1)}_{:n:} W_Q^t, \qquad
K_n = H^{(l-1)}_{:n:} W_K^t, \qquad
V_n = H^{(l-1)}_{:n:} W_V^t,
$
where $W_Q^t, W_K^t, W_V^t \in \mathbb{R}^{d_h \times d_h}$ are learnable. The attention weights are
$
A_n = \mathrm{softmax}\!\left(\frac{Q_n K_n^\top}{\sqrt{d_h}}\right),
$
and the attended representation is
$
\widetilde{H}_{:n:} = A_n V_n.
$
A feed-forward refinement with residual connection then gives
$
Z_{:n:} = \mathrm{FFN}(\widetilde{H}_{:n:}) + \widetilde{H}_{:n:}.
$
Stacking over all nodes yields
$
Z = \mathcal{T}_1\!\left(H^{(l-1)}\right) \in \mathbb{R}^{T_h \times N \times d_h}.
$
This module captures both local variations and long-range temporal dependencies for each sensor.

\subsubsection{Mamba-based Spatial Propagation}

Given $Z$, the spatial module propagates information across sensors at each time step. For a fixed time index $\tau$, let
$
Z_{\tau::} \in \mathbb{R}^{N \times d_h}
$
denote the sensor-wise feature matrix. FAST treats the sensor dimension as the propagation axis and applies a Mamba-style selective state-space operator along this dimension.

To capture both local and long-range spatial cues, the spatial module uses two branches before output projection. An input projection branch maps features into the state-space propagation space:
$
U_{\tau} = Z_{\tau::} W_u,
$
where $W_u \in \mathbb{R}^{d_h \times d_h}$ is learnable. A local convolutional branch extracts short-range spatial patterns:
$
G_{\tau} = \mathrm{Conv1D}(Z_{\tau::}),
$
where the convolution operates along the sensor dimension. The projected sequence is then processed by selective state-space propagation:
$
S_{\tau} = \mathcal{S}(U_{\tau}),
$
where $\mathcal{S}(\cdot)$ denotes the Mamba-based selective scan, whose complexity is linear in $N$.

The two branches are fused by gating and output projection:
$
H_{\tau::}^{s} = \left( \mathrm{SiLU}(G_{\tau}) \odot S_{\tau} \right) W_o,
$
where $\odot$ is the Hadamard product and $W_o \in \mathbb{R}^{d_h \times d_h}$ is learnable. Applying this operator to all time steps gives
$
H^{s} = \Phi(Z) \in \mathbb{R}^{T_h \times N \times d_h}.
$

This design avoids an explicit adjacency matrix. Instead, FAST learns spatial interactions through selective state-space propagation, while the local convolution branch preserves short-range locality cues, enabling efficient modeling of long-range inter-sensor dependencies on large sensor networks.

\subsubsection{Temporal Refinement and Block Output}

After spatial propagation, FAST applies a second temporal module to refine temporal dynamics under the updated spatial context:
$
H^{(l)} = \mathcal{T}_2\!\left(H^{s}\right).
$
The second temporal module has the same form as the first but operates on spatially updated features. Thus the TST block is
$
H^{(l)} = \mathcal{T}_2^{(l)}\!\left(\Phi^{(l)}\!\left(\mathcal{T}_1^{(l)}\!\left(H^{(l-1)}\right)\right)\right).
$
By alternating temporal reasoning and spatial propagation, the block lets temporal evolution be shaped by spatial context and, in turn, spatial interactions be informed by temporal features. This role-specialized design uses attention for flexible temporal interactions and Mamba for linear-time spatial propagation over sensors.

\subsection{Multi-Source Spatiotemporal Embedding}

Before the TST blocks, FAST maps heterogeneous inputs into a unified latent space. The embedding layer integrates four sources of information: traffic observations, temporal context, node identity, and temporal position.

\paragraph{Traffic Flow Encoding.}
Raw traffic observations are projected into a latent representation by a nonlinear multilayer projection:
$
X_{\text{data}} = \mathrm{MLP}(X),
$
where $X_{\text{data}} \in \mathbb{R}^{T_h \times N \times d}$.

\paragraph{Temporal Context Encoding.}
To model periodic traffic patterns, we use two learnable embedding tables:
$
E_{\text{day}} \in \mathbb{R}^{7 \times d}, \qquad
E_{\text{time}} \in \mathbb{R}^{288 \times d},
$
for day-of-week and intra-day time slots. The resulting embeddings $X_{\text{day}}$ and $X_{\text{time}}$ encode daily and weekly periodicities.

\paragraph{Joint Node-Position Encoding.}
To represent node-specific temporal behaviors, we introduce a learnable joint spatiotemporal embedding
$
E_{\text{pn}} \in \mathbb{R}^{T_h \times N \times d},
$
which jointly encodes node identity and temporal position. This allows FAST to model node-specific temporal patterns without using absolute timestamps.

\paragraph{Embedding Fusion.}
The final embedded input is obtained by concatenating all components:
$
X_{\text{emb}} = \mathrm{Concat}(X_{\text{data}}, X_{\text{day}}, X_{\text{time}}, X_{\text{pn}}),
$
where
$
X_{\text{emb}} \in \mathbb{R}^{T_h \times N \times d_h}, \qquad d_h = 4d.
$
This unified representation is the input to the stacked TST blocks.

\subsection{Hierarchical Skip Prediction Head}

To exploit representations from different depths, FAST uses skip connections from all TST blocks to the final predictor. Let
$
H_l \in \mathbb{R}^{T' \times N \times d}
$
denote the output of the $l$-th block. Each block contributes to the final prediction through a learnable projection:
$
\hat{Y} = \sum_{l=1}^{L} H_l W_l,
$
where $W_l \in \mathbb{R}^{d \times T_{\text{pred}}}$ is trainable.

This multi-level skip prediction lets shallow blocks preserve short-term patterns while deeper blocks contribute more abstract long-range dependencies. Such hierarchical fusion improves feature reuse, stabilizes optimization, and benefits multi-step forecasting.

\subsection{Complexity Analysis}

We analyze one TST block in terms of historical length $T_h$, number of sensors $N$, hidden dimension $d_h$, and state dimension $d_s$ of the spatial state-space module.

\paragraph{Temporal module.}
For each node, temporal self-attention computes pairwise interactions across $T_h$ time steps with complexity
$
\mathcal{O}(T_h^2 d_h).
$
Applied to all $N$ nodes, one temporal module costs
$
\mathcal{O}(N T_h^2 d_h).
$

\paragraph{Spatial module.}
The local convolution branch scales linearly with the sensor dimension, and the Mamba-based selective state-space propagation is also linear in $N$. Applied across all $T_h$ time steps, the spatial module costs
$
\mathcal{O}(T_h N d_h d_s).
$

\paragraph{Overall complexity.}
A single TST block contains two temporal modules and one spatial module, so its total complexity is
$
\mathcal{O}(N T_h^2 d_h + T_h N d_h d_s),
$
up to constant factors. The key advantage is that FAST retains expressive temporal attention while avoiding quadratic spatial complexity in the number of sensors. In particular, the spatial cost grows linearly with $N$, making FAST suitable for large-scale traffic networks.

\subsection{Training Objective}

FAST is trained for multi-step forecasting with the mean squared error between ground truth $Y$ and prediction $\hat{Y}$:
$
\mathcal{L}_{\text{MSE}} = \frac{1}{T_{\text{pred}} \cdot N} \|Y - \hat{Y}\|_F^2,
$
where $\|\cdot\|_F$ denotes the Frobenius norm.

\noindent
Overall, FAST integrates role-specialized temporal attention, Mamba-based spatial propagation, multi-source spatiotemporal embedding, and hierarchical skip prediction within a unified framework for scalable traffic forecasting.
\section{Experiments}

We evaluate \textbf{FAST} from two perspectives: 
(1) its forecasting accuracy against representative baselines on standard traffic benchmarks, and 
(2) the contributions of the proposed multi-source embedding and TST design.

\subsection{Experimental Protocol}

\subsubsection{Datasets}
We conduct experiments on three widely used traffic forecasting benchmarks: \textbf{PeMS04}, \textbf{PeMS07}, and \textbf{PeMS08}. These datasets are collected from inductive loop detectors on California highway systems, with traffic flow recorded every 5 minutes. They differ in spatial scale and traffic dynamics, providing a representative testbed for spatiotemporal forecasting.

\subsubsection{Task and Metrics}
We consider the standard multi-step forecasting setting. All methods take the most recent 12 time steps (the previous hour) as input and predict the next 12 time steps (the next hour). We report \textbf{MAE}, \textbf{RMSE}, and \textbf{MAPE}.

\subsubsection{Baseline Models}
We compare \textbf{FAST} with 10 representative baselines from four families: \textbf{iTransformer}, \textbf{PatchTST}, \textbf{STGCN} \cite{STGCN}, \textbf{DCRNN} \cite{DCRNN}, \textbf{GWNET} \cite{GWNET}, \textbf{STG-NCDE} \cite{STGNCDE}, \textbf{GMAN} \cite{GMAN}, \textbf{ST-WA} \cite{ST-WA}, \textbf{MCST-Mamba} \cite{MCST-Mamba}, and \textbf{TSMamba} \cite{TSMamba}.

\subsubsection{Implementation Details}
All experiments are conducted on an NVIDIA GeForce RTX 3080 Ti GPU (12 GB). \textbf{FAST} is implemented in PyTorch 2.0.0. Computational resources supporting this study were provided in part by Shanghai Elegant Technology Co., Ltd. Each dataset is split into training, validation, and test sets with a ratio of 8:1:1. 

The hidden dimension \(d\) is selected from \(\{32, 64\}\), and the number of ST-Blocks from \(\{1, 2, 3\}\). Each temporal attention module uses 4 or 8 heads. In the spatial Mamba block, the state expansion factor is 64, the local convolution width is 2, and the block expansion factor is 1. We use a dropout rate of 0.1. Training uses AdamW with learning rate 0.001 and batch size 32 for up to 100 epochs. Early stopping with patience 5 is applied, and the final model is selected by validation performance.

\subsection{Main Results}

\begin{table*}[htbp]
\centering
\caption{Overall forecasting performance on PeMS04, PeMS07, and PeMS08. The best result is in \textbf{bold}, and the second-best result is \underline{underlined}.}
\resizebox{\linewidth}{!}{
\begin{tabular}{cc cc cccc cc cc c}
\toprule
\multirow{2}{*}{Dataset} & \multirow{2}{*}{Metric} 
& \multicolumn{2}{c}{Time Series Forecasting} 
& \multicolumn{4}{c}{Graph Neural Network-Based} 
& \multicolumn{2}{c}{Attention-Based} 
& \multicolumn{2}{c}{Mamba-Based} 
& Ours \\
\cmidrule(lr){3-4} \cmidrule(lr){5-8} \cmidrule(lr){9-10} \cmidrule(lr){11-12} \cmidrule(lr){13-13}
& & iTransformer & PatchTST & STGCN & DCRNN & GWNET & STGNCDE & GMAN & ST-WA & MCST-Mamba & TSMamba & \textbf{FAST} \\
\midrule
\multirow{3}{*}{\centering PeMS04} 
& MAE       & 22.56 & 22.29 & 21.77 & 22.73 & 19.37 & 19.22 & 19.12 & \underline{19.07} & 20.13 & 19.08 & \textbf{19.00} \\
& MAPE (\%) & 16.16 & 16.34 & 13.88 & 14.76 & 13.29 & \underline{12.78} & 13.18 & \textbf{12.53} & 13.00 & 12.88 & 13.44 \\
& RMSE      & 35.22 & 33.70 & 34.76 & 36.56 & 31.73 & 31.07 & 31.62 & \underline{31.03} & 33.23 & 34.03 & \textbf{30.94} \\
\midrule
\multirow{3}{*}{\centering PeMS07} 
& MAE       & 24.60 & 23.94 & 22.91 & 23.62 & 21.24 & \underline{20.60} & 20.96 & 20.73 & 21.02 & 20.67 & \textbf{20.50} \\
& MAPE (\%) & 11.11 & 13.52 & 11.97 & 12.29 & 9.10 & 8.87 & 9.06 & \underline{8.75} & 8.80 & 8.89 & \textbf{8.63} \\
& RMSE      & 37.82 & 34.40 & \textbf{33.46} & 36.52 & 34.11 & 34.02 & 34.09 & 34.07 & 33.75 & 33.70 & \underline{33.58} \\
\midrule
\multirow{3}{*}{\centering PeMS08} 
& MAE       & 20.04 & 19.10 & 17.86 & 18.18 & \underline{15.08} & 15.45 & 15.33 & 15.40 & 16.44 & 16.23 & \textbf{14.66} \\
& MAPE (\%) & 12.28 & 12.63 & 11.22 & 11.23 & \textbf{9.52} & 9.91 & 10.12 & 9.96 & 10.00 & 9.91 & \underline{9.68} \\
& RMSE      & 31.88 & 25.80 & 27.13 & 28.17 & 24.87 & 24.80 & 24.91 & \underline{24.64} & 24.54 & 24.71 & \textbf{23.59} \\
\bottomrule
\end{tabular}
}
\label{tab:prediction_results}
\vspace{-6pt}
\end{table*}

Table~\ref{tab:prediction_results} summarizes the overall results. \textbf{FAST} achieves the best MAE on all three datasets, the best RMSE on PeMS04 and PeMS08, and the best MAPE on PeMS07. Across the 9 metric-dataset pairs, \textbf{FAST} ranks first on 6 and second on 2, showing strong and consistent performance.

The gains are most pronounced on \textbf{PeMS08}, where \textbf{FAST} improves MAE from 15.08 to 14.66 over the strongest competing baseline and reduces RMSE from 24.54 to 23.59. Compared with time-series, graph-based, attention-based, and Mamba-based baselines, \textbf{FAST} more effectively combines expressive temporal reasoning and efficient spatial interaction modeling within a unified TST block.

\subsection{Ablation Studies}

\subsubsection{Effect of the Multi-Source Spatio-Temporal Embedding}

\begin{table}[htbp]
\centering
\caption{Ablation results of the multi-source spatio-temporal embedding.}
\footnotesize
\begin{tabular}{c cccc}
\toprule
Metric & Model & PeMS04 & PeMS07 & PeMS08 \\
\midrule
\multirow{2}{*}{MAE} 
& w/o Embedding & 23.45 & 25.21 & 19.98 \\
& \textbf{FAST} & \textbf{19.00} & \textbf{20.50} & \textbf{14.66} \\
\midrule
\multirow{2}{*}{MAPE (\%)} 
& w/o Embedding & 16.12 & 11.29 & 12.87 \\
& \textbf{FAST} & \textbf{13.44} & \textbf{8.63} & \textbf{9.68} \\
\midrule
\multirow{2}{*}{RMSE} 
& w/o Embedding & 36.84 & 39.03 & 30.76 \\
& \textbf{FAST} & \textbf{30.94} & \textbf{33.58} & \textbf{23.59} \\
\bottomrule
\end{tabular}
\label{tab:ablation1}
\vspace{-6pt}
\end{table}

Table~\ref{tab:ablation1} evaluates the contribution of the proposed multi-source spatio-temporal embedding. Removing the embedding causes consistent degradation on all datasets and metrics. For example, on PeMS08, removing the embedding increases MAE from 14.66 to 19.98 and RMSE from 23.59 to 30.76, showing that jointly encoding traffic observations, temporal context, and node-specific information is crucial for accurate spatiotemporal representation learning.

\subsubsection{Effect of the Temporal-Spatial-Temporal Block Design}

To examine the role of the proposed TST design, we compare \textbf{FAST} with three variants:
\begin{itemize}
    \item \textit{w/ Attention}: replacing the spatial Mamba module with attention;
    \item \textit{w/ Mamba}: replacing temporal attention with Mamba for temporal modeling;
    \item \textit{Swapped Mamba-Attention}: swapping the order of the Mamba and attention modules.
\end{itemize}

\begin{figure}[tbh!]
    \centering
    \includegraphics[scale=.14]{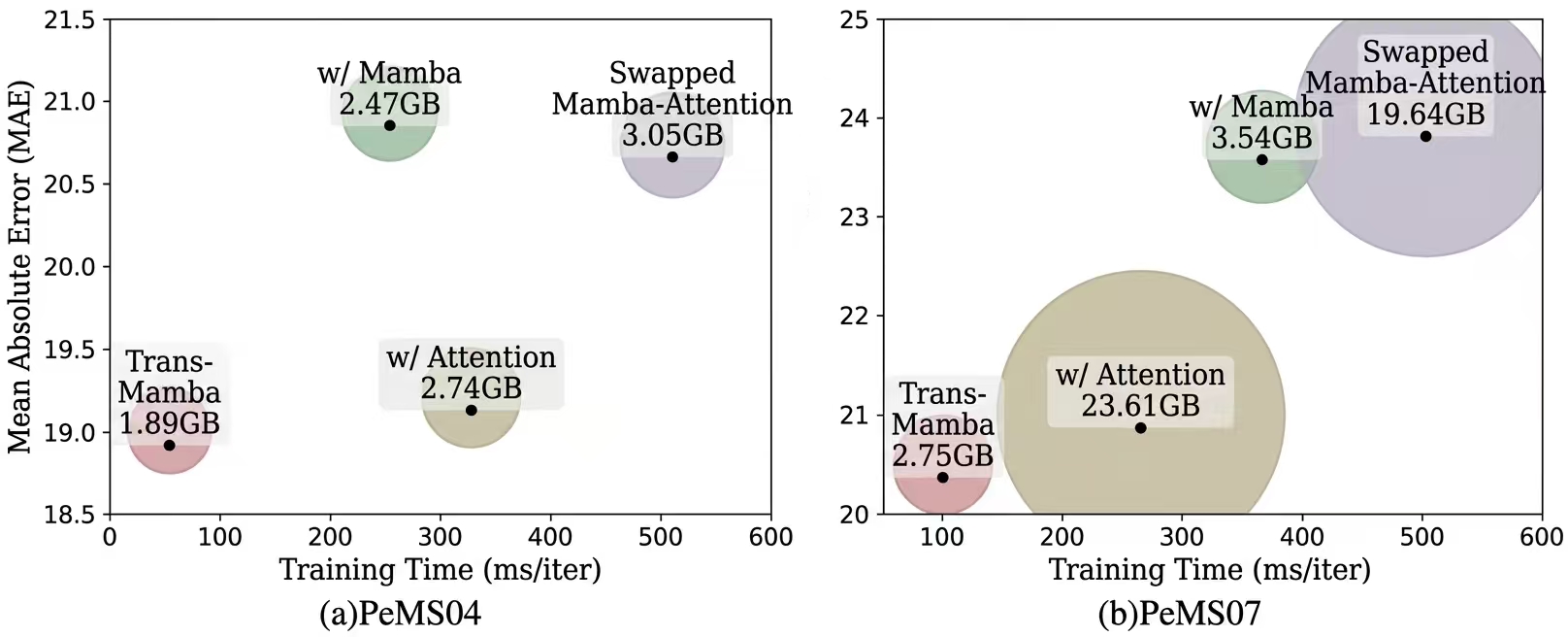}
    \caption{Ablation of the spatiotemporal block design on PeMS04 and PeMS07. Bubble position indicates MAE and training time, while bubble size denotes GPU memory usage. FAST achieves the best trade-off among accuracy, efficiency, and memory compared with \textit{w/ Attention}, \textit{w/ Mamba}, and \textit{Swapped Mamba-Attention}.}
    \label{fig:ablation}
\end{figure}

Fig.~\ref{fig:ablation} compares the four designs in terms of MAE, training time, and GPU memory usage. \textbf{FAST} achieves the best overall trade-off among accuracy, efficiency, and memory consumption. Replacing the spatial Mamba module with attention increases computational cost, replacing temporal attention with Mamba degrades temporal modeling quality, and swapping the module order also hurts performance. This shows that the gain comes not only from combining attention and Mamba, but from assigning them to appropriate roles within the TST structure.

Overall, the ablation results verify that both key components of \textbf{FAST}---the multi-source embedding and the TST spatiotemporal block---are essential to its final performance.

\section{Conclusion}

In this work, we addressed a central challenge in spatiotemporal traffic forecasting: an effective model must simultaneously capture long-range temporal dependencies, non-local spatial interactions, and computational scalability over large sensor networks. To tackle this challenge, we proposed \textbf{FAST}, a unified framework that combines attention and state-space modeling for traffic prediction. Specifically, FAST uses attention-based modules for expressive temporal reasoning and a Mamba-based state-space module for efficient spatial propagation.  n addition, FAST incorporates a learnable multi-source spatiotemporal embedding to encode heterogeneous traffic observations and contextual information, together with a multi-level skip prediction mechanism to aggregate hierarchical representations for stable multi-step forecasting. Extensive experiments on PeMS04, PeMS07, and PeMS08 demonstrate the effectiveness of the proposed framework. FAST achieves the best MAE and RMSE on all three benchmarks against strong baselines from GNN-, attention-, Transformer-, and Mamba-based families, while maintaining competitive MAPE. In particular, on PeMS08, FAST reduces RMSE by up to 4.3\% and MAE by up to 2.8\% relative to the strongest baseline. Ablation studies further show that both the proposed embedding strategy and the TST block contribute substantially to the final performance, supporting the overall design of FAST. As future work, it would be valuable to extend FAST to more challenging settings, such as multivariate traffic forecasting, adaptive spatial structure modeling, and multitask urban prediction with richer contextual signals.

\bibliographystyle{IEEEtran}

\ifarxiv
\bibliography{reference,reference_arxiv}
\else
\bibliography{reference}

@phdthesis{HA,
  author = {B. L. Smith},
  title  = {Forecasting freeway traffic flow for intelligent transportation systems application},
  school = {University of Virginia},
  year   = {1995}
}

@article{ARIMA,
  author  = {S. V. Kumar and L. Vanajakshi},
  title   = {Short-term traffic flow prediction using seasonal {ARIMA} model with limited input data},
  journal = {European Transport Research Review},
  volume  = {7},
  pages   = {1--9},
  year    = {2015}
}

@article{SVR,
  author  = {C.-H. Wu and J.-M. Ho and D.-T. Lee},
  title   = {Travel-time prediction with support vector regression},
  journal = {IEEE Transactions on Intelligent Transportation Systems},
  volume  = {5},
  number  = {4},
  pages   = {276--281},
  year    = {2004}
}

@article{LR,
  author  = {D. Li},
  title   = {Predicting short-term traffic flow in urban based on multivariate linear regression model},
  journal = {Journal of Intelligent \& Fuzzy Systems},
  volume  = {39},
  number  = {2},
  pages   = {1417--1427},
  year    = {2020}
}

@article{GWNET,
  author         = {Z. Wu and S. Pan and G. Long and J. Jiang and C. Zhang},
  title          = {Graph WaveNet for Deep Spatial-Temporal Graph Modeling},
  journal        = {arXiv},
  eprint         = {1906.00121},
  archivePrefix  = {arXiv},
  primaryClass   = {cs.LG},
  year           = {2019}
}

@inproceedings{DCRNN,
  author    = {Y. Li and R. Yu and C. Shahabi and Y. Liu},
  title     = {Diffusion Convolutional Recurrent Neural Network: Data-Driven Traffic Forecasting},
  booktitle = {International Conference on Learning Representations (ICLR)},
  year      = {2018}
}

@article{STGCN,
  author         = {B. Yu and H. Yin and Z. Zhu},
  title          = {Spatio-Temporal Graph Convolutional Networks: A Deep Learning Framework for Traffic Forecasting},
  journal        = {arXiv},
  eprint         = {1709.04875},
  archivePrefix  = {arXiv},
  primaryClass   = {cs.LG},
  year           = {2017}
}

@inproceedings{STFGNN,
  author    = {M. Li and Z. Zhu},
  title     = {Spatial-Temporal Fusion Graph Neural Networks for Traffic Flow Forecasting},
  booktitle = {Proceedings of the AAAI Conference on Artificial Intelligence},
  volume    = {35},
  pages     = {4189--4196},
  year      = {2021}
}

@inproceedings{AGCRN,
  author    = {L. Bai and L. Yao and C. Li and X. Wang and C. Wang},
  title     = {Adaptive Graph Convolutional Recurrent Network for Traffic Forecasting},
  booktitle = {Advances in Neural Information Processing Systems (NeurIPS)},
  volume    = {33},
  pages     = {17804--17815},
  year      = {2020}
}

@inproceedings{GMAN,
  author    = {C. Zheng and X. Fan and C. Wang and J. Qi},
  title     = {GMAN: A Graph Multi-Attention Network for Traffic Prediction},
  booktitle = {Proceedings of the AAAI Conference on Artificial Intelligence},
  volume    = {34},
  pages     = {1234--1241},
  year      = {2020}
}

@inproceedings{ST-WA,
  author    = {R.-G. Cirstea and B. Yang and C. Guo and T. Kieu and S. Pan},
  title     = {Towards Spatio-Temporal Aware Traffic Time Series Forecasting},
  booktitle = {Proceedings of the 2022 IEEE 38th International Conference on Data Engineering (ICDE)},
  pages     = {2900--2913},
  year      = {2022}
}

@article{PatchTST,
  author         = {Y. Nie and N. H. Nguyen and P. Sinthong and J. Kalagnanam},
  title          = {A Time Series is Worth 64 Words: Long-Term Forecasting with Transformers},
  journal        = {arXiv},
  eprint         = {2211.14730},
  archivePrefix  = {arXiv},
  primaryClass   = {cs.LG},
  year           = {2022}
}

@article{iTransformer,
  author         = {Y. Liu and T. Hu and H. Zhang and H. Wu and S. Wang and L. Ma and M. Long},
  title          = {iTransformer: Inverted Transformers are Effective for Time Series Forecasting},
  journal        = {arXiv},
  eprint         = {2310.06625},
  archivePrefix  = {arXiv},
  primaryClass   = {cs.LG},
  year           = {2023}
}

@inproceedings{STGNCDE,
  author    = {J. Choi and H. Choi and J. Hwang and N. Park},
  title     = {Graph Neural Controlled Differential Equations for Traffic Forecasting},
  booktitle = {Proceedings of the AAAI Conference on Artificial Intelligence},
  volume    = {36},
  pages     = {6367--6374},
  year      = {2022}
}

@inproceedings{HIPPO,
  author    = {A. Gu and T. Dao and S. Ermon and A. Rudra and C. R{\'e}},
  title     = {Hippo: Recurrent Memory with Optimal Polynomial Projections},
  booktitle = {Advances in Neural Information Processing Systems (NeurIPS)},
  volume    = {33},
  pages     = {1474--1487},
  year      = {2020}
}

@article{S4,
  author         = {A. Gu and K. Goel and C. R{\'e}},
  title          = {Efficiently Modeling Long Sequences with Structured State Spaces},
  journal        = {arXiv},
  eprint         = {2111.00396},
  archivePrefix  = {arXiv},
  primaryClass   = {cs.LG},
  year           = {2021}
}

@article{Mamba,
  author         = {A. Gu and T. Dao},
  title          = {Mamba: Linear-Time Sequence Modeling with Selective State Spaces},
  journal        = {arXiv},
  eprint         = {2312.00752},
  archivePrefix  = {arXiv},
  primaryClass   = {cs.LG},
  year           = {2023}
}

@article{U-Mamba,
  author         = {J. Ma and F. Li and B. Wang},
  title          = {U-Mamba: Enhancing Long-Range Dependency for Biomedical Image Segmentation},
  journal        = {arXiv},
  eprint         = {2401.04722},
  archivePrefix  = {arXiv},
  primaryClass   = {cs.CV},
  year           = {2024}
}

@article{VIM,
  author         = {L. Zhu and B. Liao and Q. Zhang and X. Wang and W. Liu and X. Wang},
  title          = {Vision Mamba: Efficient Visual Representation Learning with Bidirectional State Space Model},
  journal        = {arXiv},
  eprint         = {2401.09417},
  archivePrefix  = {arXiv},
  primaryClass   = {cs.CV},
  year           = {2024}
}

@article{VideoMamba,
  author         = {K. Li and X. Li and Y. Wang and Y. He and Y. Wang and L. Wang and Y. Qiao},
  title          = {VideoMamba: State Space Model for Efficient Video Understanding},
  journal        = {arXiv},
  eprint         = {2403.06977},
  archivePrefix  = {arXiv},
  primaryClass   = {cs.CV},
  year           = {2024}
}

@article{TSMamba,
  author         = {H. Ma and Y. Chen and W. Zhao and J. Yang and Y. Ji and X. Xu and X. Liu and H. Jing and S. Liu and G. Yang},
  title          = {A Mamba Foundation Model for Time Series Forecasting},
  journal        = {arXiv},
  eprint         = {2411.02941},
  archivePrefix  = {arXiv},
  year           = {2024}
}

@article{MCST-Mamba,
  author         = {M. Hamad and M. Mabrok and N. Zorba},
  title          = {MCST-Mamba: Multivariate Mamba-Based Model for Traffic Prediction},
  journal        = {arXiv},
  eprint         = {2507.03927},
  archivePrefix  = {arXiv},
  year           = {2025}
}

@inproceedings{Transformer,
  author    = {Ashish Vaswani and Noam Shazeer and Niki Parmar and Jakob Uszkoreit and Llion Jones and Aidan N. Gomez and Lukasz Kaiser and Illia Polosukhin},
  title     = {Attention Is All You Need},
  booktitle = {Advances in Neural Information Processing Systems (NeurIPS)},
  pages     = {5998--6008},
  year      = {2017}
}

@ARTICLE{11059994,
  author={Duan, Wenchang and Gao, Zhenguo and He, Jiwan and Xian, Jinguo},
  journal={IEEE Transactions on Intelligent Transportation Systems}, 
  title={Bayesian Critique-Tune-based Reinforcement Learning With Adaptive Pressure for Multi-Intersection Traffic Signal Control}, 
  year={2025},
  volume={26},
  number={10},
  pages={14968-14983},
  doi={10.1109/TITS.2025.3581858}
}

@inproceedings{duanadaptive,
  title={Adaptive Context Length Optimization with Low-Frequency Truncation for Multi-Agent Reinforcement Learning},
  author={Duan, Wenchang and Yu, Yaoliang and He, Jiwan and Shi, Yi},
  booktitle={The Thirty-ninth Annual Conference on Neural Information Processing Systems}
}

@inproceedings{liu2025efficient,
  title={Efficient partitioning vision transformer on edge devices for distributed inference},
  author={Liu, Xiang and Song, Yijun and Li, Xia and Sun, Yifei and Lan, Huiying and Liu, Zemin and Jiang, Linshan and Li, Jialin},
  booktitle={2025 IEEE 45th International Conference on Distributed Computing Systems (ICDCS)},
  pages={286--296},
  year={2025},
  organization={IEEE}
}

@article{liu2024fedlpa,
  title={Fedlpa: One-shot federated learning with layer-wise posterior aggregation},
  author={Liu, Xiang and Liu, Liangxi and Ye, Feiyang and Shen, Yunheng and Li, Xia and Jiang, Linshan and Li, Jialin},
  journal={Advances in Neural Information Processing Systems},
  volume={37},
  pages={81510--81548},
  year={2024}
}

@article{liu2025one,
  title={One-shot federated learning methods: A practical guide},
  author={Liu, Xiang and Tang, Zhenheng and Li, Xia and Song, Yijun and Ji, Sijie and Liu, Zemin and Han, Bo and Jiang, Linshan and Li, Jialin},
  journal={arXiv preprint arXiv:2502.09104},
  year={2025}
}

@article{ouyang2024learn,
  title={Learn from global correlations: Enhancing evolutionary algorithm via spectral gnn},
  author={Ouyang, Kaichen and Ke, Zong and Fu, Shengwei and Liu, Lingjie and Zhao, Puning and Hu, Dayu},
  journal={arXiv preprint arXiv:2412.17629},
  year={2024}
}

@article{ke2025stable,
  title={A stable technical feature with GRU-CNN-GA fusion},
  author={Ke, Zong and Shen, Jiaqing and Zhao, Xuanyi and Fu, Xinghao and Wang, Yang and Li, Zichao and Liu, Lingjie and Mu, Huailing},
  journal={Applied Soft Computing},
  pages={114302},
  year={2025},
  publisher={Elsevier}
}

@inproceedings{INTENT,
  title={INTENT: Invariance and Discrimination-aware Noise Mitigation for Robust Composed Image Retrieval},
  author={Chen, Zhiwei and Hu, Yupeng and Fu, Zhiheng and Li, Zixu and Huang, Jiale and Huang, Qinlei and Wei, Yinwei},
  booktitle={Proceedings of the AAAI Conference on Artificial Intelligence},
  volume={40},
  number={25},
  pages={20463--20471},
  year={2026}
}

@inproceedings{HUD, 
  title = {HUD: Hierarchical Uncertainty-Aware Disambiguation Network for Composed Video Retrieval}, 
  author = {Chen, Zhiwei and Hu, Yupeng and Li, Zixu and Fu, Zhiheng and Wen, Haokun and Guan, Weili}, 
  booktitle = {Proceedings of the ACM International Conference on Multimedia}, 
  pages = {6143–6152}, 
  year = {2025} 
}

@article{REFINE,
  title={REFINE: Composed Video Retrieval via Shared and Differential Semantics Enhancement},
  author={Hu, Yupeng and Li, Zixu and Chen, Zhiwei and Huang, Qinlei and Fu, Zhiheng and Xu, Mingzhu and Nie, Liqiang},
  journal={ACM Transactions on Multimedia Computing, Communications and Applications},
  year={2026},
  publisher={ACM New York, NY}
}

@inproceedings{liu2023regformer,
  title={RegFormer: an efficient projection-aware transformer network for large-scale point cloud registration},
  author={Liu, Jiuming and Wang, Guangming and Liu, Zhe and Jiang, Chaokang and Pollefeys, Marc and Wang, Hesheng},
  booktitle={Proceedings of the IEEE/CVF International Conference on Computer Vision},
  pages={8451--8460},
  year={2023}
}

@inproceedings{liu2024difflow3d,
  title={DifFlow3D: Toward Robust Uncertainty-Aware Scene Flow Estimation with Iterative Diffusion-Based Refinement},
  author={Liu, Jiuming and Wang, Guangming and Ye, Weicai and Jiang, Chaokang and Han, Jinru and Liu, Zhe and Zhang, Guofeng and Du, Dalong and Wang, Hesheng},
  booktitle={Proceedings of the IEEE/CVF Conference on Computer Vision and Pattern Recognition},
  pages={15109--15119},
  year={2024}
}

@inproceedings{liu2024dvlo,
  title={Dvlo: Deep visual-lidar odometry with local-to-global feature fusion and bi-directional structure alignment},
  author={Liu, Jiuming and Zhuo, Dong and Feng, Zhiheng and Zhu, Siting and Peng, Chensheng and Liu, Zhe and Wang, Hesheng},
  booktitle={European Conference on Computer Vision},
  pages={475--493},
  year={2024},
  organization={Springer}
}

@inproceedings{jiang2025stg,
  title={STG-Avatar: Animatable Human Avatars via Spacetime Gaussian},
  author={Jiang, Guangan and Zhang, Tianzi and Li, Dong and Zhao, Zhenjun and Li, Haoang and Li, Mingrui and Wang, Hongyu},
  booktitle={2025 IEEE/RSJ International Conference on Intelligent Robots and Systems (IROS)},
  pages={20058--20065},
  year={2025},
  organization={IEEE}
}

@article{chan2026adagar,
  title={AdaGaR: Adaptive Gabor Representation for Dynamic Scene Reconstruction},
  author={Chan, Jiewen and Zhao, Zhenjun and Liu, Yu-Lun},
  journal={arXiv preprint arXiv:2601.00796},
  year={2026}
}

@inproceedings{yan2025turboreg,
  title={Turboreg: Turboclique for robust and efficient point cloud registration},
  author={Yan, Shaocheng and Shi, Pengcheng and Zhao, Zhenjun and Wang, Kaixin and Cao, Kuang and Wu, Ji and Li, Jiayuan},
  booktitle={Proceedings of the IEEE/CVF International Conference on Computer Vision},
  pages={26371--26381},
  year={2025}
}

@article{wu2025spatiotemporal,
  title={Spatiotemporal multi-view continual dictionary learning with graph diffusion},
  author={Wu, Sheng and Zhang, Jinlai},
  journal={Knowledge-Based Systems},
  volume={316},
  pages={113388},
  year={2025},
  publisher={Elsevier}
}

@article{ning2025multi,
  title={Multi-Resolution Context Augmentation and Dual Channel Attention for 3D lane detection},
  author={Ning, Qirui and Zhang, Jinlai and Xie, Yuhang and Liu, Kaifeng and Gao, Kai and Chen, Bin and Chen, Gengbiao and Fan, Qing and Liu, Hui and Du, Ronghua},
  journal={IEEE Internet of Things Journal},
  year={2025},
  publisher={IEEE}
}

@inproceedings{liu2025research,
  title={Research and practice of advertisement recommendation algorithm based on graph neural network},
  author={Liu, Kuangcong and Yang, Shini and Xia, Jiayi},
  booktitle={Proceedings of the 2nd International Symposium on Integrated Circuit Design and Integrated Systems},
  pages={210--215},
  year={2025}
}

@inproceedings{zhao2025efficient,
  title={Efficient Cold-Start Recommendation via BPE Token-Level Embedding Initialization with LLM},
  author={Zhao, Yushang and Han, Xinyue and Leng, Qian ox and Sun, Qianyi and Lyu, Haotian and Zhou, Chengrui},
  booktitle={2025 3rd International Conference on Artificial Intelligence and Automation Control (AIAC)},
  pages={194--198},
  year={2025},
  organization={IEEE}
}

@misc{lai2026transformers,
  author       = {Lai, Longying and Cheng, Zhiyuan and Cheng, Kai and Qi, Xiaoxi},
  title        = {Do Transformers Always Win? An Empirical Study of Semantic Embeddings for Short-Text E-commerce Reviews},
  year         = {2026},
  month        = mar,
  doi          = {10.21203/rs.3.rs-9163424/v1},
  url          = {https://doi.org/10.21203/rs.3.rs-9163424/v1},
  note         = {Research Square preprint, Version 1, posted 20 March 2026}
}

@misc{cheng2026regime,
  author       = {Cheng, Kai and Qi, Xiaoxi and Cheng, Zhiyuan and Lai, Longying and Liu, Xuan},
  title        = {Regime-Dependent Volatility Dynamics: Evidence from Time-Series Analysis},
  year         = {2026},
  month        = jan,
  doi          = {10.2139/ssrn.6321958},
  url          = {https://dx.doi.org/10.2139/ssrn.6321958},
  note         = {SSRN working paper. Date Written: January 24, 2026; posted March 11, 2026}
}

@misc{chen2025autoneuralcodesigningvisionlanguagemodels,
  title={AutoNeural: Co-Designing Vision-Language Models for NPU Inference}, 
  author={Wei Chen and Liangmin Wu and Yunhai Hu and Zhiyuan Li and Zhiyuan Cheng and Yicheng Qian and Lingyue Zhu and Zhipeng Hu and Luoyi Liang and Qiang Tang and Zhen Liu and Han Yang},
  year={2025},
  eprint={2512.02924},
  archivePrefix={arXiv},
  primaryClass={cs.CL},
  url={https://arxiv.org/abs/2512.02924}, 
}

@article{codes2026,
doi = {10.20944/preprints202603.1152.v1},
url = {https://doi.org/10.20944/preprints202603.1152.v1},
year = {2026},
month = {March},
author = {Lan Hu and Yuting Xin and Binqi Shen and Hanyu Cai and Lier Jin},
title = {CoDES: A Context-Efficient Framework for Enhancing Small Language Models via Domain-Specific Adaptation and Model Ensembling},
journal = {Preprints},
eprint = {202603.1152}
}

@article{zhou2023fastpillars,
  title={FastPillars: A Deployment-friendly Pillar-based 3D Detector},
  author={Zhou, Sifan and Zhang, Xinyu and Chu, Xiangxiang and Zhang, Bo and Zhao, Ziyu and Lu, Xiaobo},
  journal={IEEE Transactions on Circuits and Systems for Video Technology}, 
  year={2025},
  doi={10.1109/TCSVT.2025.3633725}
}

@inproceedings{anonymous2025comptrack,
  title={CompTrack: Information Bottleneck-Guided Low-Rank Dynamic Token Compression for Point Cloud Tracking}, 
  author={Sifan Zhou and Yichao Cao and Jiahao Nie and Yuqian Fu and Ziyu Zhao and Xiaobo Lu and Shuo Wang},
  booktitle={The Fortieth AAAI Conference on Artificial Intelligence}, 
  year={2025}, 
  url={https://openreview.net/forum?id=nXExYROmVe}
}

@inproceedings{gsq,
    title = "{GSQ}-Tuning: Group-Shared Exponents Integer in Fully Quantized Training for {LLM}s On-Device Fine-tuning",
    author = "Zhou, Sifan and Wang, Shuo and Yuan, Zhihang and Shi, Mingjia and Shang, Yuzhang and Yang, Dawei",
    booktitle = "Findings of the Association for Computational Linguistics: ACL 2025",
    month = jul,
    year = "2025",
    address = "Vienna, Austria",
    publisher = "Association for Computational Linguistics",
    pages = "22971--22988",
    ISBN = "979-8-89176-256-5",
}

@article{wu2025efficiency,
  title={From Efficiency to Adaptivity: A Deeper Look at Adaptive Reasoning in Large Language Models},
  author={Wu, Chao and Li, Baoheng and Gao, Mingchen and Wang, Zhenyi},
  journal={arXiv preprint arXiv:2511.10788},
  year={2025}
}

@inproceedings{luan2025dynamic,
  title={Dynamic neural fortresses: An adaptive shield for model extraction defense},
  author={Luan, Siyu and Wang, Zhenyi and Shen, Li and Gu, Zonghua and Wu, Chao and Tao, Dacheng},
  booktitle={The Thirteenth International Conference on Learning Representations},
  year={2025}
}

@inproceedings{wang-etal-2025-filter,
    title = "Filter-And-Refine: A {MLLM} Based Cascade System for Industrial-Scale Video Content Moderation",
    author = "Wang, Zixuan and Shi, Jinghao and Liang, Hanzhong and Shen, Xiang and Wen, Vera and Chen, Zhiqian and Wu, Yifan and Zhang, Zhixin and Xiong, Hongyu",
    booktitle = "Proceedings of the 63rd Annual Meeting of the Association for Computational Linguistics (Volume 6: Industry Track)",
    year = "2025",
    pages = "873--880",
    doi = "10.18653/v1/2025.acl-industry.62"
}

@inproceedings{wang-etal-2025-reasoning-enhanced,
    title = "Reasoning-Enhanced Domain-Adaptive Pretraining of Multimodal Large Language Models for Short Video Content Governance",
    author = "Wang, Zixuan and Sun, Yu and Wang, Hongwei and Jing, Baoyu and Shen, Xiang and Dong, Xin and Hao, Zhuolin and Xiong, Hongyu and Song, Yang",
    booktitle = "Proceedings of the 2025 Conference on Empirical Methods in Natural Language Processing: Industry Track",
    year = "2025",
    pages = "1104--1112",
    doi = "10.18653/v1/2025.emnlp-industry.77"
}

@misc{yu2026rulesfallshortagentdriven,
      title={When Rules Fall Short: Agent-Driven Discovery of Emerging Content Issues in Short Video Platforms}, 
      author={Chenghui Yu and Hongwei Wang and Junwen Chen and Zixuan Wang and Bingfeng Deng and Zhuolin Hao and Hongyu Xiong and Yang Song},
      year={2026},
      eprint={2601.11634},
      archivePrefix={arXiv},
      primaryClass={cs.CV}
}

@article{li2025efficient,
  title={An Efficient Solution Method for Solving Convex Separable Quadratic Optimization Problems},
  author={Li, Shaoze and Wu, Junhao and Lu, Cheng and Deng, Zhibin and Fang, Shu-Cherng},
  journal={arXiv preprint arXiv:2510.11554},
  year={2025}
}

@article{wu2023semidefinite,
  title={A semidefinite relaxation based global algorithm for two-level graph partition problem.},
  author={Wu, Junhao and Lu, Cheng and Li, Shaoze and Deng, Zhibin},
  journal={Journal of Industrial \& Management Optimization},
  volume={19},
  number={9},
  year={2023}
}

@article{he2021towards,
  title={Towards cleaner heating production in rural areas: Identifying optimal regional renewable systems with a case in Ningxia, China},
  author={He, Jiaming and Wu, Yunna and Wu, Junhao and Li, Shaoze and Liu, Fangtong and Zhou, Jianli and Liao, Mingjuan},
  journal={Sustainable Cities and Society},
  volume={75},
  pages={103288},
  year={2021},
  publisher={Elsevier}
}

@article{zeng2025janusvln,
  title={Janusvln: Decoupling semantics and spatiality with dual implicit memory for vision-language navigation},
  author={Zeng, Shuang and Qi, Dekang and Chang, Xinyuan and Xiong, Feng and Xie, Shichao and Wu, Xiaolong and Liang, Shiyi and Xu, Mu and Wei, Xing},
  journal={arXiv preprint arXiv:2509.22548},
  year={2025}
}

@article{zeng2025futuresightdrive,
  title={Futuresightdrive: Thinking visually with spatio-temporal cot for autonomous driving},
  author={Zeng, Shuang and Chang, Xinyuan and Xie, Mengwei and Liu, Xinran and Bai, Yifan and Pan, Zheng and Xu, Mu and Wei, Xing and Guo, Ning},
  journal={arXiv preprint arXiv:2505.17685},
  year={2025}
}

@inproceedings{zeng2026priordrive,
  title={PriorDrive: Enhancing Online HD Mapping with Unified Vector Priors},
  author={Zeng, Shuang and Chang, Xinyuan and Liu, Xinran and Yuan, Yujian and Liang, Shiyi and Pan, Zheng and Xu, Mu and Wei, Xing},
  booktitle={Proceedings of the AAAI Conference on Artificial Intelligence},
  volume={40},
  number={15},
  pages={12313--12321},
  year={2026}
}

@ARTICLE{Feng_2024_NoiseBox,
author = {Feng, Chen and Tzimiropoulos, Georgios and Patras, Ioannis},
journal = {IEEE Transactions on Circuits and Systems for Video Technology},
title = {{NoiseBox: Towards More Efficient and Effective Learning with Noisy Labels}},
year = {2024},
doi = {10.1109/TCSVT.2024.3426994},
month = {7}
}

@inproceedings{chen2025prosac,
title = {{PROSAC}: Provably Safe Certification for Machine Learning Models under Adversarial Attacks},
author = {Feng, Chen and Liu, Ziquan and Zhi, Zhuo and Bogunovic, Ilija and Gerner-Beuerle, Carsten and Rodrigues, Miguel.R.D},
booktitle = {The 39th Annual AAAI Conference on Artificial Intelligence (AAAI) [Oral]},
year = {2025},
doi = {10.1609/aaai.v39i3.32300}
}

@inproceedings{feng2025noisy,
title = {Noisy but Valid: Robust Statistical Evaluation of {LLM}s with Imperfect Judges},
author = {Feng, Chen and Shen, Minghe and Balashankar, Ananth and Gerner-Beuerle, Carsten and Rodrigues, Miguel.R.D},
booktitle = {The Fourteenth International Conference on Learning Representations (ICLR)},
year = {2026},
url = {https://openreview.net/forum?id=hEhxreaLdU}
}

@article{JUNJIE2025581,
  title = {YOLOv8-DDS: A lightweight model based on pruning and distillation for early detection of root mold in barley seedling},
  journal = {Information Processing in Agriculture},
  volume = {12},
  number = {4},
  pages = {581-594},
  year = {2025},
  author = {Huang, Junjie and Ma, Zheng and Wu, Yuzhu and Bao, Yujian and Wang, Yizhe and Su, Zhongbin and Guo, Lifeng},
  doi = {https://doi.org/10.1016/j.inpa.2025.07.004}
}

@article{pang2026voice,
  title={From Voice to Safety: Language AI Powered Pilot-ATC Communication Understanding for Airport Surface Movement Collision Risk Assessment},
  author={Pang, Yutian and Kendall, Andrew Paul and Porcayo, Alex and Barsotti, Mariah and Jain, Anahita and Clarke, John-Paul},
  journal={Transportation Research Part C: Emerging Technologies},
  volume={184},
  pages={105540},
  year={2026},
  publisher={Elsevier}
}

@article{hu2024decentralized,
  title={Decentralized Graph-Based Multi-Agent Reinforcement Learning Using Reward Machines},
  author={Hu, Jueming and Xu, Zhe and Wang, Weichang and Qu, Guannan and Pang, Yutian and Liu, Yongming},
  journal={Neurocomputing},
  volume={564},
  pages={126974},
  year={2024},
  publisher={Elsevier}
}

@article{hu2025reinforcement,
  title={Reinforcement Learning Driven Integrated Detection and Mitigation of UAV GPS Spoofing Attacks},
  author={Hu, Jueming and Ammar, Mohammad and Hussain, Bilal Zahid and Kim, Jaewon and Khan, Irfan},
  journal={IEEE Internet of Things Journal},
  year={2025},
  publisher={IEEE}
}

@inproceedings{deng2026rearl,
  title={{REA}-{RL}: Reflection-Aware Online Reinforcement Learning for Efficient Reasoning},
  author={Hexuan Deng and Wenxiang Jiao and Xuebo Liu and Jun Rao and Min Zhang},
  booktitle={The Fourteenth International Conference on Learning Representations (ICLR)},
  year={2026},
  url={https://openreview.net/forum?id=E6keG5QDct}
}

@article{pesf-kd,
  author    = {Jun Rao and Xv Meng and Liang Ding and Shuhan Qi and Xuebo Liu and Min Zhang and Dacheng Tao},
  title     = {Parameter-Efficient and Student-Friendly Knowledge Distillation},
  journal   = {{IEEE} Trans. Multim.},
  pages     = {1--12},
  year      = {2023},
  doi       = {10.1109/TMM.2023.3321480}
}

@inproceedings{rao2025dynamicsamplingadaptsiterative,
  title={Dynamic Sampling that Adapts: Self-Aware Iterative Data Persistent Optimization for Mathematical Reasoning}, 
  author={Jun Rao and Xuebo Liu and Hexuan Deng and Zepeng Lin and Zixiong Yu and Jiansheng Wei and Xiaojun Meng and Min Zhang},
  year={2026},
  booktitle={Findings of the Association for Computational Linguistics: ACL 2026},
  url={https://arxiv.org/abs/2505.16176}
}

@article{yu2026spatiotemporal,
  title={Spatiotemporal Alignment for Remote Sensing Image Recovery via Terrain-Aware Diffusion},
  author={Yu, Zhenyu and Jiang, Haoran and Wang, Pei and Lin, Zizhen and Xiang, Yong},
  journal={ICASSP 2026},
  year={2026},
  publisher={Preprints}
}

@article{meng2026tri,
  title={Tri-Subspaces Disentanglement for Multimodal Sentiment Analysis},
  author={Meng, Chunlei and Luo, Jiabin and Yan, Zhenglin and Yu, Zhenyu and Fu, Rong and Gan, Zhongxue and Ouyang, Chun},
  journal={CVPR 2026},
  year={2026},
  publisher={arXiv preprint arXiv:2602.19585}
}

@inproceedings{yu2025cotextor,
  title={CoTextor: Training-Free Modular Multilingual Text Editing via Layered Disentanglement and Depth-Aware Fusion},
  author={Yu, Zhenyu and Idris, Mohd Yamani Idna and Wang, Pei and Qureshi, Rizwan},
  booktitle={NeurIPS 2025},
  year={2025}
}

@article{li2025fully,
  title={A fully data-driven approach for realistic traffic signal control using offline reinforcement learning},
  author={Li, Jianxiong and Lin, Shichao pagan and Shi, Tianyu and Tian, Chujie and Mei, Yu and Song, Jian and Zhan, Xianyuan and Li, Ruimin},
  journal={Data Science for Transportation},
  volume={7},
  number={3},
  pages={1--18},
  year={2025},
  publisher={Springer Nature Singapore}
}

@article{zhang2025ccma,
  title={Ccma: A framework for cascading cooperative multi-agent in autonomous driving merging using large language models},
  author={Zhang, Miao and Fang, Zhenlong and Wang, Tianyi and Lu, Shuai and Wang, Xueqian and Shi, Tianyu},
  journal={Expert Systems with Applications},
  volume={282},
  pages={127717},
  year={2025},
  publisher={Pergamon}
}

@article{xiao2026echo,
  title={ECHO-2: A Large Scale Distributed Rollout Framework for Cost-efficient Reinforcement Learning},
  author={Xiao, Jie and Chen, Meng and Ren, Qingnan and Jingwei, Song and Huang, Jiaqi and Deng, Yangshen and Tong, Chris and Chen, Wanyi and Wang, Suli and Bi, Ziqian and others},
  journal={arXiv preprint arXiv:2602.02192},
  year={2026}
}

@inproceedings{luo-etal-2025-dynamicner,
    title = "{D}ynamic{NER}: A Dynamic, Multilingual, and Fine-Grained Dataset for {LLM}-based Named Entity Recognition",
    author = "Luo, Hanjun and Jin, Yingbin and Wang, Yiran and Li, Xinfeng and Shang, Tong and Liu, Xuecheng and Chen, Ruizhe and Wang, Kun and Salam, Hanan and Wen, Qingsong and Liu, Zuozhu",
    booktitle = "Proceedings of the 2025 Conference on Empirical Methods in Natural Language Processing",
    year = "2025",
    pages = "16511--16535",
    doi = "10.18653/v1/2025.emnlp-main.835"
}

@misc{luo2026agentauditorhumanlevelsafetysecurity,
    title={AgentAuditor: Human-Level Safety and Security Evaluation for LLM Agents}, 
    author={Hanjun Luo and Shenyu Dai and Chiming Ni and Xinfeng Li and Guibin Zhang and Kun Wang and Tongliang Liu and Hanan Salam},
    year={2026},
    eprint={2506.00641},
    archivePrefix={arXiv},
    primaryClass={cs.AI}
}

@article{ji2025finestate,
  title={FineState-Bench: A Comprehensive Benchmark for Fine-Grained State Control in GUI Agents},
  author={Ji, Fengxian and Yang, Jingpu and Song, Zirui and Wang, Yuanxi and Cui, Zhexuan and Li, Yuke and Jiang, Qian and Fang, Miao and Chen, Xiuying},
  journal={arXiv preprint arXiv:2508.09241},
  year={2025}
}

@article{jiang2026magma,
  title={MAGMA: A Multi-Graph based Agentic Memory Architecture for AI Agents},
  author={Jiang, Dongming and Li, Yi and Li, Guanpeng and Li, Bingzhe},
  journal={arXiv preprint arXiv:2601.03236},
  year={2026}
}

@article{jiang2026anatomy,
  title={Anatomy of Agentic Memory: Taxonomy and Empirical Analysis of Evaluation and System Limitations},
  author={Jiang, Dongming and Li, Yi and Wei, Songtao and Yang, Jinxin and Kishore, Ayushi and Zhao, Alysa and Kang, Dingyi and Hu, Xu and Chen, Feng and Li, Qiannan and others},
  journal={arXiv preprint arXiv:2602.19320},
  year={2026}
}
\fi

\end{document}